\documentclass[sigconf]{acmart}

\copyrightyear{2026}
\acmYear{2026}
\setcopyright{cc}
\setcctype{by}

\acmConference[MM '26]
{Proceedings of the 34th ACM International Conference on Multimedia}
{November 10--14, 2026}
{Rio de Janeiro, Brazil}

\acmBooktitle{Proceedings of the 34th ACM International Conference on Multimedia
(MM '26), November 10--14, 2026, Rio de Janeiro, Brazil}

\acmDOI{10.1145/3767308.3836325}
\acmISBN{979-8-4007-2213-4/2026/11}

\graphicspath{{figures/}}

\newcommand{\R}{\mathbb{R}}
\DeclareMathOperator*{\argmax}{arg\,max}

\title[Embodied Multimodal Grounding via Semantic 3DGS]
{Embodied Multimodal Grounding for Open-Vocabulary Mobile Manipulation via Semantic 3D Gaussian Splatting}

\author{Huosen Ou}
\orcid{0009-0002-3313-5107}
\affiliation{%
  \institution{The Hong Kong University of Science and Technology (Guangzhou)}
  \city{Guangzhou}
  \state{Guangdong}
  \country{China}
}
\email{hou061@connect.hkust-gz.edu.cn}

\author{Dongni Song}
\orcid{0009-0003-5614-7861}
\affiliation{%
  \institution{The Hong Kong University of Science and Technology (Guangzhou)}
  \city{Guangzhou}
  \state{Guangdong}
  \country{China}
}
\email{dsong549@connect.hkust-gz.edu.cn}

\author{Yuncong Wang}
\orcid{0009-0009-1154-3741}
\affiliation{%
  \institution{The Hong Kong University of Science and Technology (Guangzhou)}
  \city{Guangzhou}
  \state{Guangdong}
  \country{China}
}
\email{ywang321@connect.hkust-gz.edu.cn}

\author{Tao Zhou}
\correspondingauthor
\authornote{Tao Zhou and Yiding Ji are corresponding authors.}
\orcid{0009-0004-9425-5393}
\affiliation{%
  \institution{Midea Group}
  \city{Foshan}
  \state{Guangdong}
  \country{China}
}
\email{zhoutao42@midea.com}

\author{Yiding Ji}
\authornotemark[1]
\correspondingauthor
\orcid{0000-0003-2678-7051}
\affiliation{%
  \institution{The Hong Kong University of Science and Technology (Guangzhou)}
  \city{Guangzhou}
  \state{Guangdong}
  \country{China}
}
\affiliation{
  \institution{The Hong Kong University of Science and Technology}
  \city{Hong Kong}
  \country{China}
}
\email{jiyiding@hkust-gz.edu.cn}
\renewcommand{\shortauthors}{Ou et al.}

\begin{abstract}
Embodied mobile manipulation requires language, visual observations, three-dimensional scene structure, and action feasibility to be aligned before execution. We study open-vocabulary target grounding with few-shot manipulation in local household workspaces and present an embodied multimodal grounding framework that integrates active multi-view Semantic 3D Gaussian Splatting (Semantic-3DGS), reachability-aware base positioning, and a diffusion-based vision-language-action policy. A task-driven local Semantic-3DGS serves as a shared interface across active sensing, language-conditioned 3D localization, obstacle-aware scene reasoning, base preparation, and semantic conditioning of the action model. To preserve pretrained action priors, the 3D semantic cues are injected only into the late action-expert blocks. In expanded 50-trial real-robot evaluations against representative vision-language-action (VLA) approaches, the full system achieves 60\% long-horizon success compared with 40\% for PointVLA and 28\% for DexVLA, and reaches 74\% success in heavily cluttered manipulation compared with 52\% for the single-view variant and 46\% for PointVLA. It also maintains 75\% success under a 75\,cm height shift and eliminates photo-induced false grasps. These results indicate that explicit, refreshable 3D semantic grounding can improve robustness under clutter, occlusion, viewpoint variation, and embodiment constraints.
\end{abstract}

\ccsdesc[500]{Computing methodologies~Vision for robotics}
\ccsdesc[300]{Computer systems organization~Robotic autonomy}
\ccsdesc[300]{Computing methodologies~Reconstruction}

\keywords{embodied multimedia, multimodal grounding, multimodal fusion, multimedia and language, mobile manipulation, semantic 3D Gaussian splatting, vision-language-action}

\begin{document}
\maketitle

\section{Introduction}

Recent multimodal models have substantially advanced language-guided perception and action, yet robust embodied execution in real environments remains difficult. In household mobile manipulation scenarios, a robot must identify the object referred to by a language instruction, ground it in three-dimensional space, prepare a feasible body configuration, and execute under clutter, occlusion, viewpoint variation, and embodiment constraints. This makes mobile manipulation a representative problem of \emph{embodied multimedia understanding}, where language, imagery, geometry, and robot state must be fused into grounded action.

This work focuses on two practical failure modes. First, many vision-language-action (VLA) systems still rely heavily on 2D appearance cues, making target grounding brittle under partial occlusion, clutter, or appearance-only distractors such as a tablet displaying a photo-realistic target. Second, even when the target is correctly localized, manipulation may still fail if the mobile base is poorly positioned relative to the arm workspace, particularly in long-horizon tasks or when the target height changes.

%%% rebuttal里面第一段话总结的不错，我把一些句子给加到正文里了。

We therefore propose an embodied multimodal grounding framework centered on a compact and refreshable Semantic-3DGS to address the challenge of cross-modal spatial alignment. The representation is not used only as a static 3D container: the same target-conditioned local field serves as a shared interface for active-view scoring, Gaussian-level language localization, obstacle-aware rendering, target-relative pose extraction, and semantic conditioning of the manipulation policy. We initialize geometry from VGGT~\cite{wang2025vggt}, distill CLIP/DINO semantics with mask-aware regularization inspired by Feature Splatting~\cite{qiu2024feature}, compute a reachability-aware base stance, and condition a pretrained diffusion VLA~\cite{wen2025dexvla} through late-block semantic adapters. This design aims to add explicit 3D grounding while preserving pretrained visuomotor priors.

Our scope is \emph{open-vocabulary target grounding with few-shot manipulation}, rather than zero-shot acquisition of arbitrary robot skills. Test objects are held-out instances, while the embodiment-specific manipulation policy is adapted using 10 real demonstrations per task. The local Semantic-3DGS is constructed before manipulation and refreshed when grounding fails; it is not continuously optimized inside the low-level control loop. Our Late-Block Semantic Injection is a multimodal fusion mechanism bridging heterogeneous perception and reactive control. Instead of early injection, which degrades pretrained priors, we distill DINO/CLIP features into a local Semantic-3DGS and inject dense 3D semantics only into final diffusion experts, preserving action priors while adding 3D grounding. This preserves established continuous action priors while adding robust spatial grounding. Hence, our framework is not a loose combination of existing modules: the same object-centric Semantic-3DGS is used as a shared interface for active view scoring, Gaussian-level language localization, obstacle-aware rendering, PCA semantic tokenization, target-relative pose conditioning, and zero-initialized late-block adaptation.

\begin{figure*}[t]
    \centering
    \includegraphics[width=\textwidth]{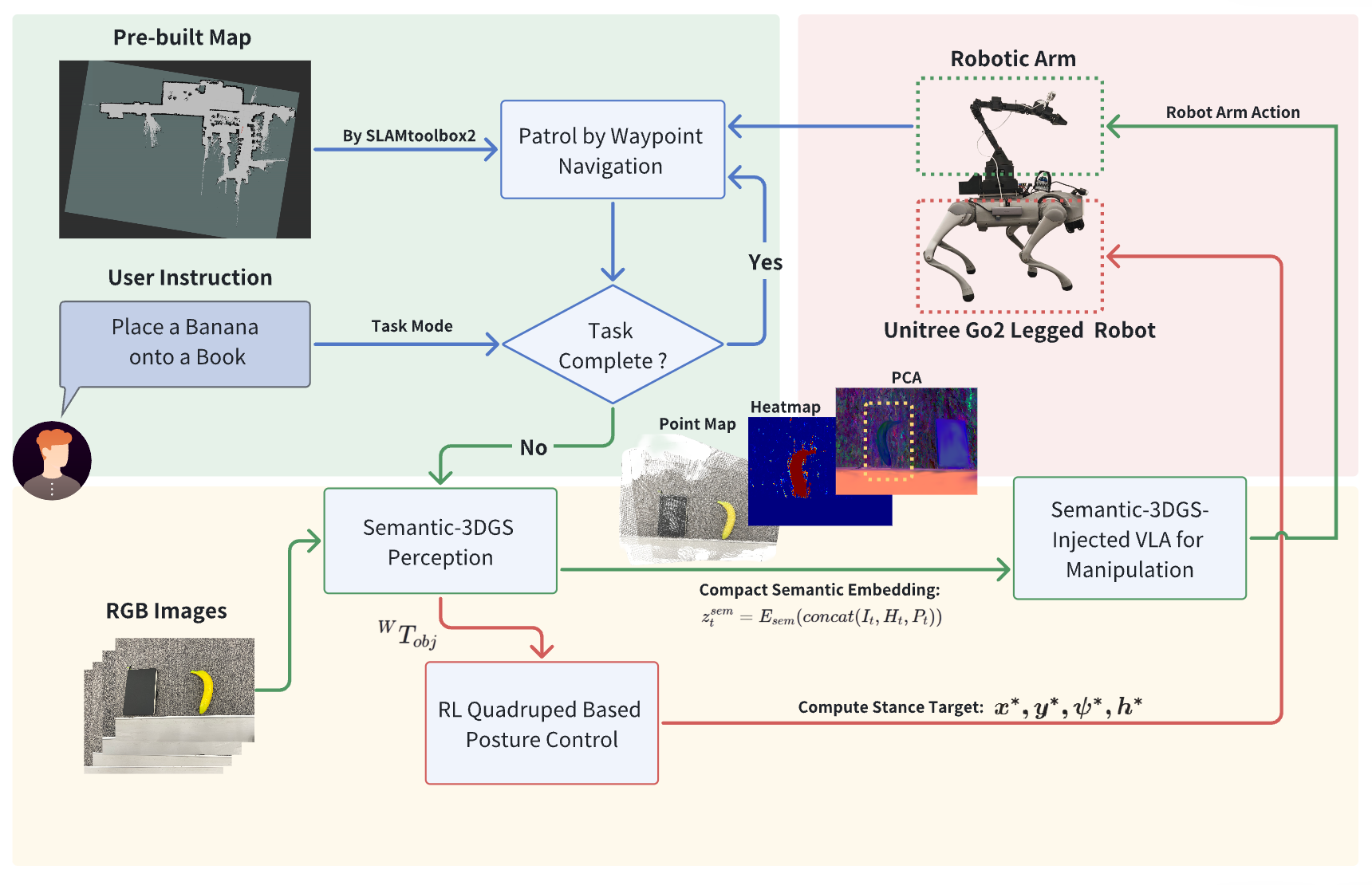}
    \caption{Overview of the proposed system. The robot patrols by waypoint navigation, switches to task mode after receiving instruction \(l\), actively acquires four local wrist-camera views, constructs a local Semantic-3DGS, estimates the target pose \({}^W T_{\mathrm{obj}}\), repositions the base to a feasible stance \((x^\star,y^\star,\psi^\star,h^\star)\), and executes a semantics-conditioned VLA manipulation policy.}
    \Description{A pipeline connecting waypoint navigation, active multi-view semantic 3D grounding, reachability-aware base preparation, and VLA manipulation. The local Semantic-3DGS supplies target localization and semantic-geometric cues to both stance selection and action generation.}
    \label{fig:pipeline}
\end{figure*}

\paragraph{Contributions.}
Our contributions are threefold. First, we formulate local-area mobile manipulation as an embodied multimodal grounding problem and construct a target-driven Semantic-3DGS from four actively acquired wrist-camera images. Second, we use this local field as a shared perception-to-action interface for active sensing, language grounding, obstacle-aware reasoning, target-relative pose extraction, and late-block conditioning of a diffusion action expert, preserving earlier pretrained action priors. Third, we couple explicit 3D grounding with reachability-aware stance preparation and validate the resulting system in real-robot few-shot manipulation under long-horizon execution, height shifts, photo deception, and heavy clutter, including expanded 50-trial studies with confidence intervals and runtime profiling.

\section{Related Work}

\subsection{Embodied Mobile Manipulation}

Embodied mobile manipulation of robots couples locomotion, perception, and contact-rich actions. Previous works have investigated quadruped manipulation, whole-body coordination, and long-horizon loco-manipulation~\cite{arm2024pedipulate,he2024learning,cheng2023legs,yu2022multi,ouyang2025long,pan2025roboduet,liu2025mlm,liu2024visual,portela2024learning,dadiotis2023whole,muramatsu2025mobile,zhang2025learning,yokoyama2023asc,zhang2024gamma, qin2026ipd}. These studies show that legged platforms extend manipulation beyond fixed tabletop settings by enabling flexible approach, posture adaptation, and operation in cluttered spaces.

Compared with static-arm manipulation, however, mobile manipulation is more sensitive to embodiment constraints such as base placement, body height, arm workspace limits, self-occlusion, and nearby obstacles. In practice, failure often stems from weak coordination between perception and action preparation rather than from the arm policy alone. Our work follows this systems perspective and connects task-conditioned 3D grounding, base positioning, and manipulation through a shared local semantic representation.

\subsection{3D Grounding and Multimodal Fusion}

A key multimedia challenge in embodied interaction is aligning language, visual observations, and geometry into a representation useful for action. Recent works have explored point-cloud-conditioned VLA models~\cite{li2025pointvla}, open-vocabulary mobile manipulation~\cite{yenamandra2023homerobot}, language-aware 3D Gaussian representations~\cite{zheng2024gaussiangrasper,qiu2024feature} and 4D Gaussian splatting for dynamic scene modeling~\cite{ou2025pose}. Compared with purely 2D grounding, 3D representations provide more stable cross-view localization and better support reasoning about target pose and nearby obstacles, especially under clutter and occlusion.

Our work aligns with this technical direction, but emphasizes a task-driven local representation rather than dense full-scene optimization or persistent global mapping. Under our settings, a refreshable Semantic-3DGS is then constructed from only four actively selected wrist-camera views and the resulting field is reused across target localization, obstacle-aware reasoning, reachability-aware stance preparation, and downstream VLA conditioning. In this way, the 3D representation serves as a common multimodal interface instead of an isolated perception output.

\subsection{Vision-Language-Action Policies}

VLA models and diffusion-based robotic manipulation policies provide scalable interfaces for embodied control~\cite{wen2025dexvla,kim2024openvla,intelligence2025pi_,chi2025diffusion,zhu2025scaling,wang2024qwen2}. Large pretrained backbones offer transferable visuomotor priors, while diffusion-based action experts improve action generation and temporal consistency. These advances have substantially improved language-conditioned manipulation.

However, when explicit 3D scene cues are absent, a policy must jointly infer object identity, spatial relation, viewpoint sufficiency, and action feasibility from limited observations. This makes end-to-end policies vulnerable to clutter, occlusion, viewpoint variation, and appearance-only distractors. Our method is complementary to this line of work: we retain a pretrained diffusion-based VLA backbone and inject explicit Semantic-3DGS cues only into late action-expert blocks. This design preserves earlier pretrained action priors while supplying target-centered 3D semantics and obstacle-aware geometry for execution.

\section{Method}
\label{sec:method}

\subsection{System Overview and Compute Split}
The full system is shown in Figure~\ref{fig:pipeline}. After receiving a language instruction \(l\), the robot performs: (1) active local multi-view observation, (2) Semantic-3DGS construction and open-vocabulary 3D localization, (3) reachability-aware base repositioning, and (4) Semantic-3DGS-conditioned VLA manipulation.

The platform of this work is a Unitree Go2 Edu quadruped equipped with standing and crouching modes, a Unitree 4D L1 LiDAR, an onboard Jetson Orin NX, and an Alicia-D 6-DoF arm with an RGB camera near the gripper. Arm joint targets, including the gripper, are streamed at 30\,Hz through ROS. Semantic-3DGS perception and VLA inference run off-board on an RTX 4090 workstation, while low-level robot control and the base posture policy run onboard. The current system targets quasi-static household manipulation rather than fast dynamic interaction.

\begin{figure}[t]
    \centering
    \includegraphics[width=\columnwidth]{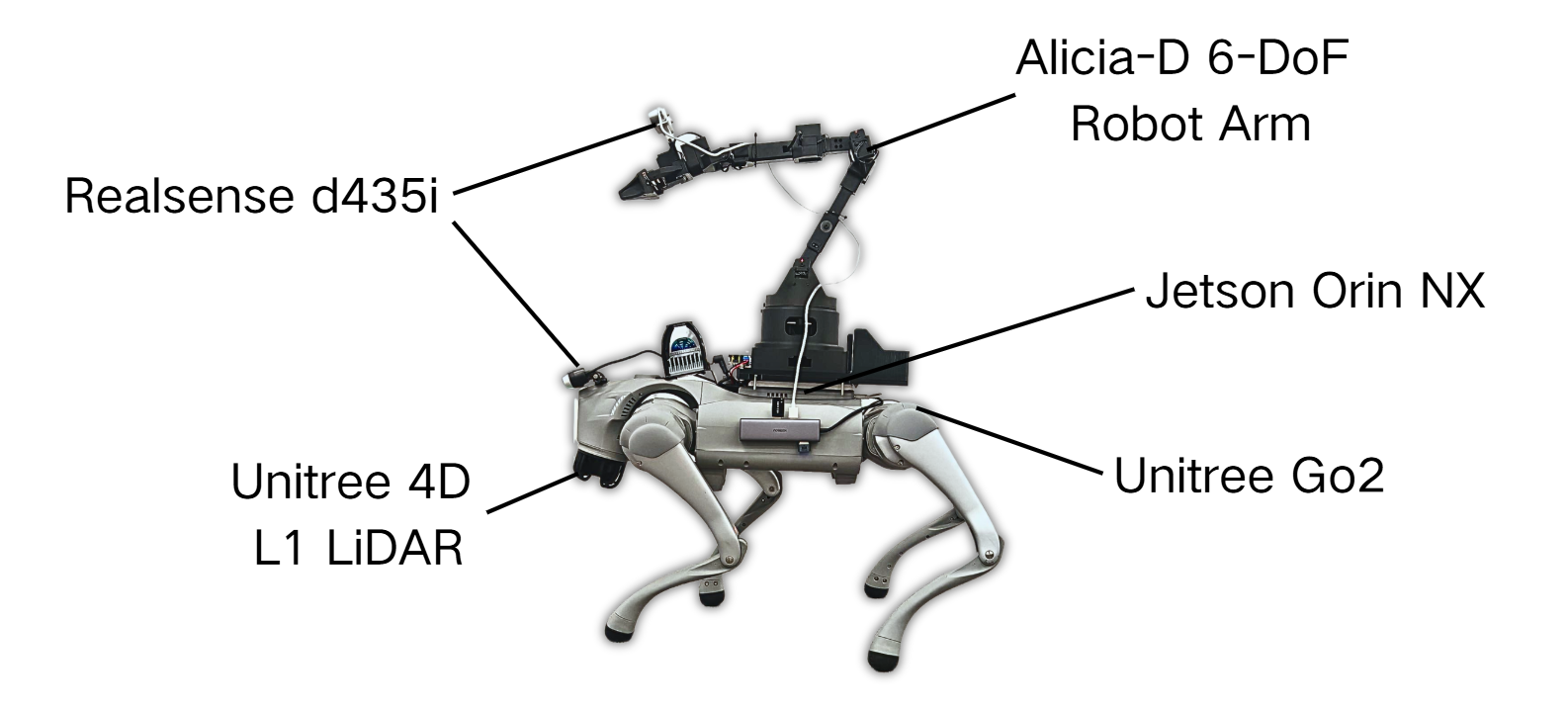}
    \caption{Robot platform: a Unitree Go2 Edu quadruped with a 6-DoF arm and an arm-mounted RGB camera for local perception and manipulation. }
    \Description{A quadruped mobile manipulator equipped with a robotic arm and an arm-mounted RGB camera for local perception and manipulation.}
    \label{fig:hardware}
\end{figure}

\subsection{Representation and Notation}
Let \(\mathcal{F}_W\) and \(\mathcal{F}_B\) denote the world and base frames, and let \({}^W T_{B,t}\in SE(3)\) be the current base pose. A multi-view image buffer is \(\mathcal{I}=\{I^{(i)}\}_{i=1}^{N}\), with \(N=4\) for the full system and \(N=1\) for \emph{Ours Single-View}. The local Semantic-3DGS is
\begin{equation}
\mathcal{G}=\{g_k\}_{k=1}^{K},\qquad
g_k=(\mu_k,\Sigma_k,c_k,\alpha_k,f_k^{C},f_k^{D}),
\end{equation}
where \(\mu_k,\Sigma_k,c_k,\alpha_k\) denote Gaussian geometry, color, and opacity, while \(f_k^C\) and \(f_k^D\) are CLIP-aligned and DINO semantic features.

\subsection{Active Multi-view Semantic-3DGS}

\begin{figure*}[t]
    \centering
    \includegraphics[width=\textwidth]{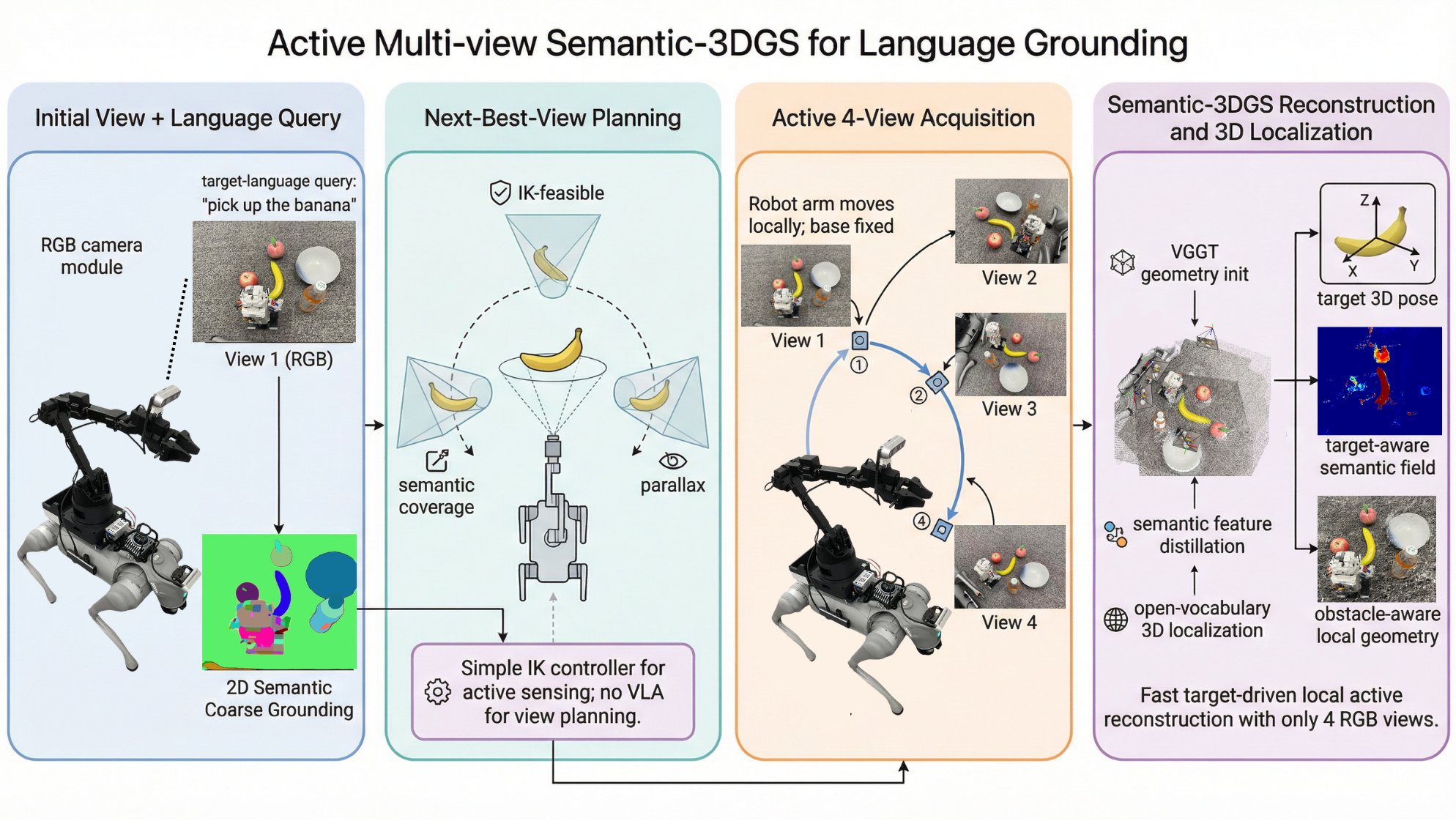}
    \caption{Active multi-view Semantic-3DGS for language grounding. Target-driven view acquisition builds a compact local semantic 3D representation from four wrist-camera images.}
    \Description{The pipeline starts from a language-conditioned initial view, scores IK-feasible candidate camera poses using target coverage, view diversity, and motion cost, captures four wrist-camera views, and reconstructs a local Semantic-3DGS for target localization and obstacle-aware geometry.}
    \label{fig:active_sem3dgs}
\end{figure*}

\subsubsection{Active View Acquisition}
From the first wrist-camera image \(I^{(1)}\), we extract a target phrase \(q(l)\), compute a language relevance map, and combine SAM masks with coarse VGGT geometry to obtain a rough 3D target support. Candidate wrist-camera poses \(\mathcal{V}_{\mathrm{cand}}\) are scored by
\begin{equation}
J(v)=\lambda_{\mathrm{cov}}C_{\mathrm{sem}}(v)
+\lambda_{\mathrm{par}}C_{\mathrm{par}}(v)
-\lambda_{\mathrm{move}}C_{\mathrm{move}}(v),
\end{equation}
where \(C_{\mathrm{sem}}\) estimates the expected semantic coverage of the coarse target support, \(C_{\mathrm{par}}\) rewards complementary viewpoints, and \(C_{\mathrm{move}}\) discourages unnecessary wrist motion. The terms are normalized before combination, and only IK-feasible candidates are retained. The next view is
\begin{equation}
v_{n+1}=\argmax_{v\in\mathcal{V}_{\mathrm{cand}}}J(v).
\end{equation}
The robot base remains fixed during sensing; a standard IK controller moves the wrist camera until four images are collected. No VLA policy is used for view planning.

\subsubsection{Geometry Initialization and Semantic Distillation}
Given \(\mathcal{I}\), a pretrained VGGT model~\cite{wang2025vggt} predicts camera parameters and dense geometry,
\begin{equation}
\{(\hat{g}^{(i)},\hat{D}^{(i)},\hat{P}^{(i)})\}_{i=1}^{N}
=\Phi_{\mathrm{VGGT}}(\mathcal{I}),
\end{equation}
which initializes the local Gaussian field. For each view, we extract CLIP and DINOv2 feature maps and SAM masks. Mask-aware average pooling refines the CLIP features, and rendered Gaussian features are optimized by cosine alignment:
\begin{equation}
\begin{split}
\mathcal{L}_{\mathrm{feat}}
=&\sum_{i,u}\left(1-\cos(\hat{F}_C^{(i)}(u),\tilde{F}_C^{(i)}(u))\right)\\
&+\lambda_D\sum_{i,u}\left(1-\cos(\hat{F}_D^{(i)}(u),F_D^{(i)}(u))\right),
\end{split}
\end{equation}
with \(\lambda_D=0.1\).

\subsubsection{Open-Vocabulary 3D Localization}
The target phrase is encoded by a frozen CLIP text encoder to obtain \(e^+\). With generic negative prompts \(\{e_j^-\}_{j=1}^{m}\), each Gaussian receives a language relevance score
\begin{equation}
s_k=\mathrm{softmax}\!\left(
\tau[\cos(f_k^C,e^+),\cos(f_k^C,e^-_1),\ldots,\cos(f_k^C,e^-_m)]
\right)_1.
\end{equation}
For the selected Gaussian support \(\mathcal{K}=\{k\mid s_k>\delta\}\), the object position is estimated by
\begin{equation}
p^W_{\mathrm{obj}}=
\frac{\sum_{k\in\mathcal{K}}w_k\mu_k}{\sum_{k\in\mathcal{K}}w_k},
\qquad
w_k=\alpha_k\,\mathrm{trace}(\Sigma_k)^{-1}.
\end{equation}
A stable 6D pose \({}^W T_{\mathrm{obj}}\) is estimated from the selected support using PCA, with optional ICP refinement when a template is available. The explicit 3D support reduces dependence on a single appearance cue and exposes both target and nearby obstacle geometry.

\subsection{Reachability-aware Base Posture Control}
After localization, the object position is transformed to the base frame:
\begin{equation}
p^B_{\mathrm{obj}}=({}^W T_{B,t})^{-1}p^W_{\mathrm{obj}}
=[x_{\mathrm{obj}},y_{\mathrm{obj}},z_{\mathrm{obj}}]^\top.
\end{equation}
We define a pre-manipulation stance
\begin{equation}
x^\star=x_{\mathrm{obj}}-d_x,\quad
y^\star=y_{\mathrm{obj}}-\mathrm{sign}(y_{\mathrm{obj}})d_y,\quad
\psi^\star=\mathrm{atan2}(y_{\mathrm{obj}},x_{\mathrm{obj}}),
\end{equation}
with \(d_x=0.35\) m and \(d_y=0.20\) m, and select the height mode by
\begin{equation}
h^\star=
\begin{cases}
h^{\mathrm{crouch}}, & z_{\mathrm{obj}}<0.30\text{ m},\\
h^{\mathrm{stand}}, & \text{otherwise}.
\end{cases}
\end{equation}
A proximal policy optimization (PPO) policy outputs leg-joint residuals and the stand/crouch switch. The controller is trained in Isaac Lab built on NVIDIA Isaac Sim with domain randomization; only the base posture policy is trained in the simulations.

\subsection{Semantic-3DGS-Conditioned VLA Manipulation Policy}

\begin{figure*}[t]
    \centering
    \includegraphics[width=\textwidth]{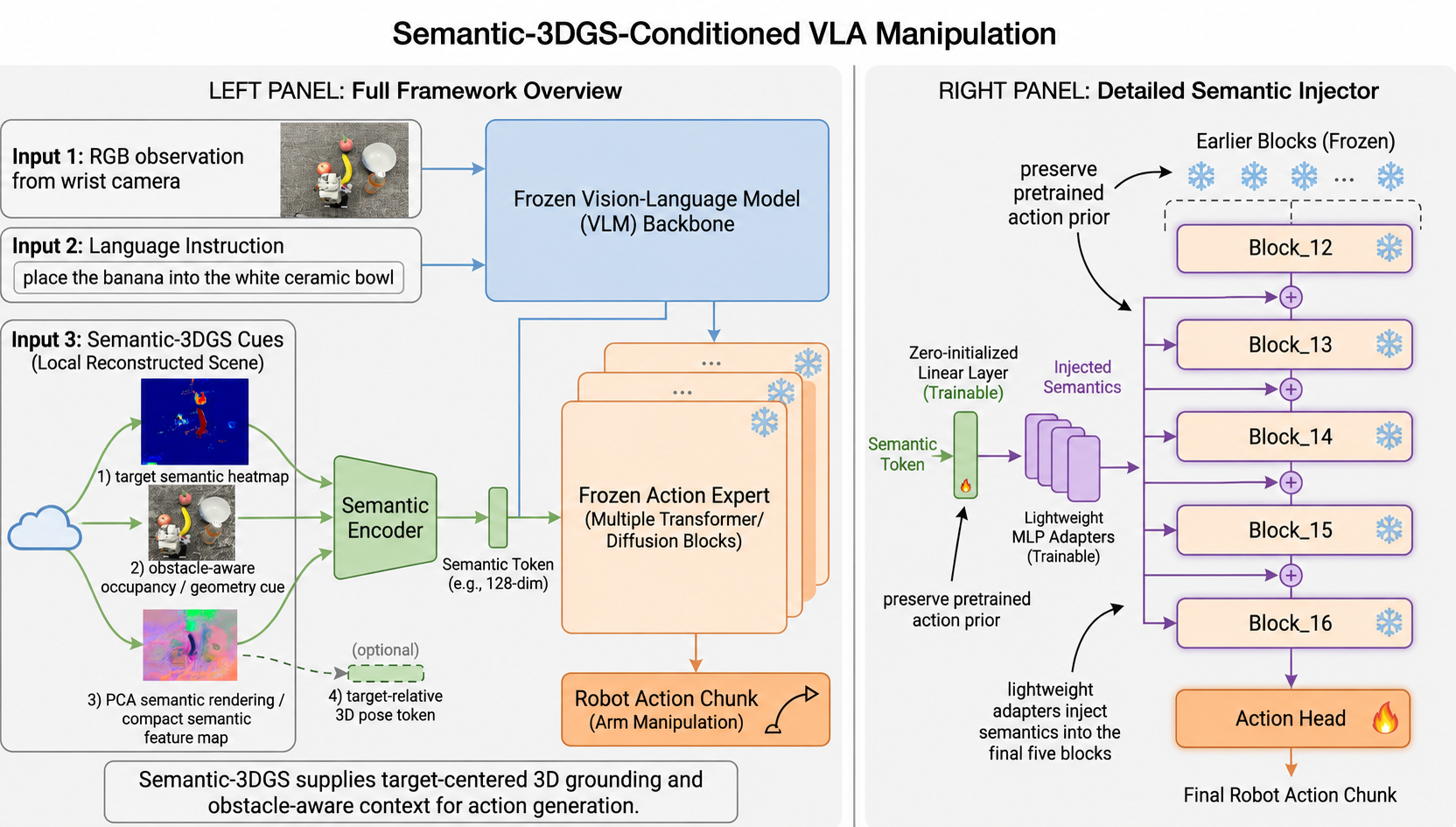}
    \caption{Semantic-3DGS-conditioned VLA manipulation. A lightweight semantic injector conditions the final five frozen action-expert blocks through trainable adapters while preserving pretrained action priors.}
    \Description{RGB observation and language are processed by the pretrained VLA, while target heatmaps, obstacle-aware geometry, a compact PCA semantic rendering, and target-relative pose from the local Semantic-3DGS are encoded into a semantic token. The token is injected through lightweight adapters into the final five action-expert blocks.}
    \label{fig:semantic_injector}
\end{figure*}

We build on a DexVLA-style policy with a Qwen2-VL backbone~\cite{wang2024qwen2} and ScaleDP action expert~\cite{zhu2025scaling}. From instruction \(l\) and \(\mathcal{G}\), we aggregate: (1) a language-conditioned target heatmap \(H_t\), (2) an obstacle-aware occupancy cue \(O_t\), (3) a three-channel PCA rendering \(P_t\) of the distilled per-Gaussian semantic field, and (4) a target-relative pose vector \(r_t=[p^B_{\mathrm{obj}},h_t]\in\R^4\). The visual cues and RGB observation are encoded as
\begin{equation}
z_t^{\mathrm{img}}=E_{\mathrm{img}}(\mathrm{concat}(I_t,H_t,O_t,P_t))\in\R^{128},
\end{equation}
and the pose cue as \(z_t^{\mathrm{pose}}=E_{\mathrm{pose}}(r_t)\in\R^{128}\). Their sum \(z_t^{\mathrm{sem}}\) is injected only into the final five diffusion blocks:
\begin{equation}
\mathcal{B}=\{L-4,L-3,L-2,L-1,L\},
\end{equation}
\begin{equation}
h_\ell\leftarrow h_\ell+A_\ell(\mathrm{Proj}(z_t^{\mathrm{sem}})),
\qquad \ell\in\mathcal{B},
\end{equation}
where \(\mathrm{Proj}\) is zero-initialized and \(A_\ell\) is a lightweight MLP adapter. The VLM backbone and all pretrained diffusion blocks remain frozen; only the semantic encoder, projection layers, late-block adapters, and embodiment-specific action head are trained. We use 10 real demonstrations per task, an action chunk of 15 steps, and two recent observation frames.

\section{Experiments}
\label{sec:experiments_main}

\subsection{Model Training, Deployment, and Evaluation Protocol}
Only the reachability-aware base posture controller is trained in Isaac Lab/Isaac Sim; the manipulation policy is trained offline from real demonstrations. Each trajectory records wrist RGB, robot proprioception, and arm joint targets at 30\,Hz. During deployment, the full system acquires a four-view RGB buffer before manipulation, whereas \emph{Ours Single-View} uses only the initial image.

Unless otherwise noted, each setting uses 30 real-robot trials with randomized initial object poses. The long-horizon and cluttered banana-to-bowl evaluations are expanded to 50 trials per method. A trial is successful only when the complete instruction is finished within the time budget (60\,s for local multi-task manipulation and 180\,s for the long-horizon task) without unsafe behavior. For the 50-trial studies, we additionally report success counts, time standard deviations, and 95\% Wilson confidence intervals.

\paragraph{Fair baseline adaptation.}
All baselines use the same robot embodiment and the same 10 teleoperated demonstrations per task. We also keep the train/test split, wrist-camera setup, action interface, navigation waypoints, base-control assistance, initial scene distributions, and evaluation budgets fixed. Each method retains its native perception representation. Thus, the comparisons mainly reflect differences in grounding and manipulation representations rather than differences in demonstrations, navigation support, or stance preparation.

\subsection{Tasks}
Next, we evaluate five groups of real-robot tasks. \emph{Few-shot multi-task manipulation} contains bottle-to-basket, banana-to-book, and ordered toy/banana/bottle placement into a white ceramic bowl. Test objects are unseen instances in task-specific fine-tuning, while semantically related object families may appear in the 10 demonstration trajectories. \emph{Long-horizon manipulation} requires opening a drawer, retrieving a banana, closing the drawer, navigating to a black chair, and placing the banana. \emph{Height adaptability} varies target platform height by 30, 60, and 75\,cm. \emph{Photo deception} replaces a real banana with a photo-realistic banana displayed on a tablet. \emph{Cluttered banana-to-bowl manipulation} places the target among multiple physical distractors and frequent occlusions.

\begin{figure*}[t]
    \centering
    \includegraphics[width=0.85\textwidth]{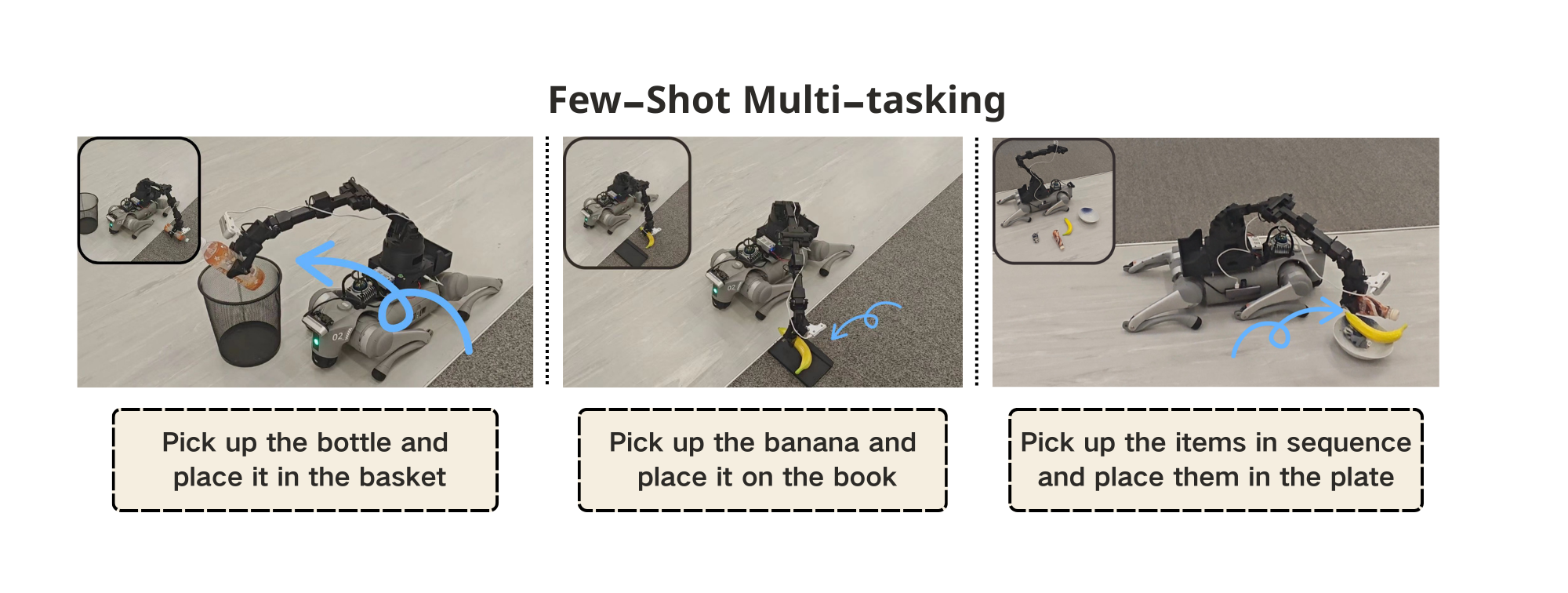}
    \caption{Few-shot multi-task manipulation scenarios. Test objects are unseen instances in task-specific fine-tuning, while semantically related object families may appear in the few-shot demonstrations.}
    \Description{Three real-robot manipulation settings involving bottle, banana, plush toy, book, basket, and a white ceramic bowl.}
    \label{fig:multitask}
\end{figure*}

\begin{figure*}[t]
    \centering
    \includegraphics[width=\textwidth]{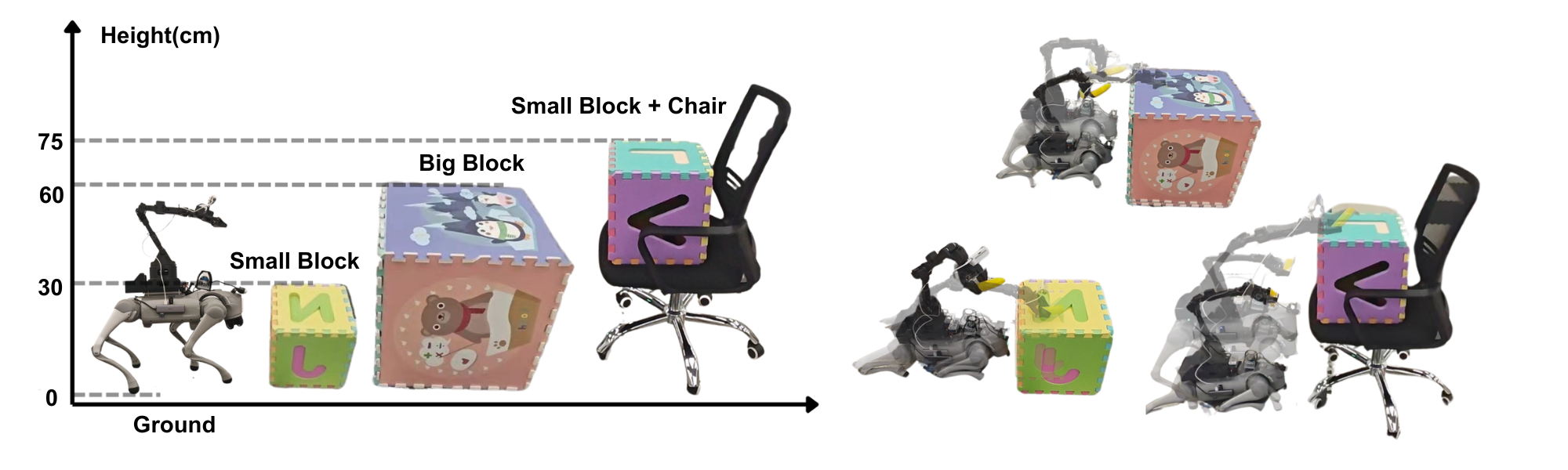}
    \caption{Height-adaptation settings. Targets are placed at different platform heights to test stance selection and reachability-aware manipulation.}
    \Description{A sequence of real-robot scenes with manipulation targets placed at increasingly different heights, illustrating the need for standing and crouching base postures.}
    \label{fig:height}
\end{figure*}

\subsection{Results}

\subsubsection{Few-shot Multi-task Manipulation}
\begin{figure}[t]
    \centering
    \includegraphics[width=1.05\columnwidth, height=7cm]{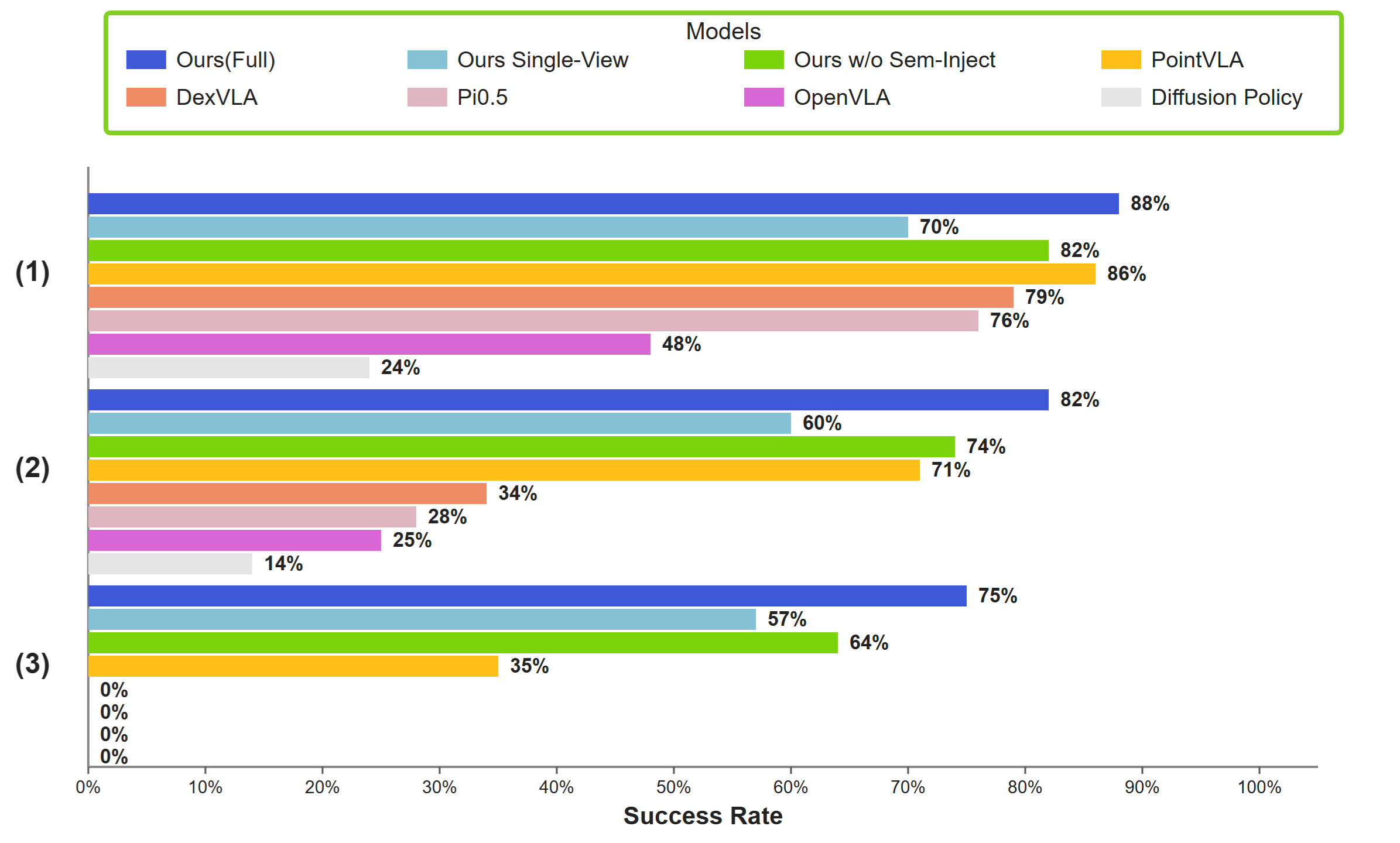}
    \vspace{-20pt}
    \caption{Success rates on the few-shot multi-task manipulation benchmark. (1) Place the banana on the book. (2) Place the bottle into the basket. (3) Place the stuffed toy, banana, and bottle into the bowl in the specified order.}
    \Description{A bar chart comparing the full method, single-view ablation, and VLA baselines across three increasingly complex manipulation tasks.}
    \label{fig:successrate}
\end{figure}

Figures~\ref{fig:multitask} and~\ref{fig:successrate} summarize the few-shot task settings and results. The full model reaches the highest average success rate (81.7\%), compared with 64.0\% for PointVLA and 37.7\% for DexVLA. The advantage becomes larger in cluttered and multi-step settings, supporting the use of explicit 3D semantic grounding beyond easy single-object cases.

\subsubsection{Long-horizon Manipulation}

\begin{figure*}[t]
    \centering
    \includegraphics[width=\textwidth]{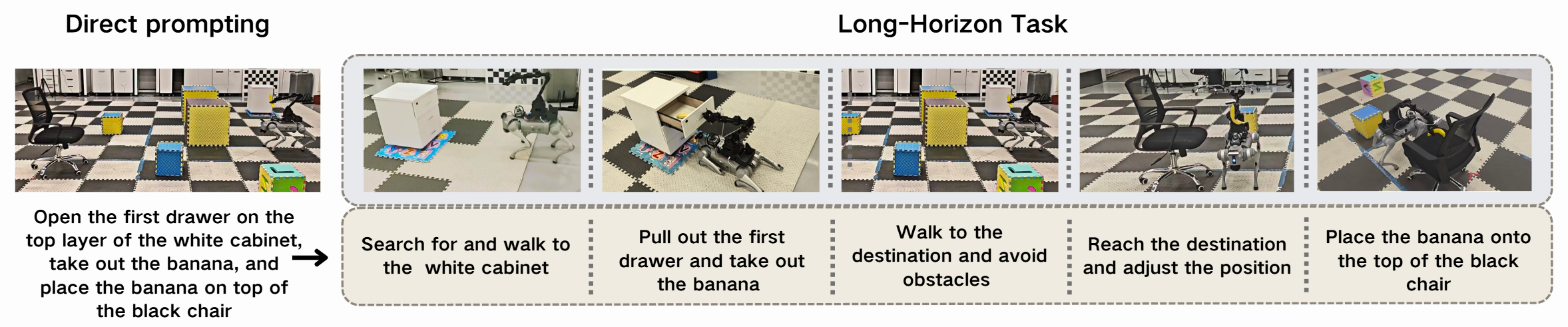}
    \caption{Long-horizon manipulation sequence. The robot opens the top drawer, retrieves the banana, closes the drawer, navigates to the target area, and places the banana on the black chair.}
    \Description{A multi-stage real-robot sequence showing drawer opening, banana retrieval, drawer closing, locomotion toward a black chair, and final object placement.}
    \label{fig:longhorizon}
\end{figure*}

Figure~\ref{fig:longhorizon} illustrates the multi-stage execution, and Table~\ref{tab:long_horizon} reports the expanded 50-trial evaluation. The complete system completes 30/50 trials (60\%) compared with 20/50 (40\%) for PointVLA and 14/50 (28\%) for DexVLA. Removing the Base-RL module reduces the success rate to 22\%, validating that mere correct grounding is insufficient when the stance is not prepared for the arm workspace.

\begin{table}[t]
\caption{Long-horizon performance over 50 trials per method. CI denotes a 95\% Wilson interval.}
\label{tab:long_horizon}
\centering
\resizebox{\columnwidth}{!}{
\begin{tabular}{lccc}
\toprule
Method & Success & 95\% CI & Avg. time (s) \\
\midrule
DexVLA & 14/50 (28\%) & [17.5, 41.7] & \(178.6\pm18.4\) \\
PointVLA & 20/50 (40\%) & [27.6, 53.8] & \(161.8\pm17.6\) \\
Ours w/o Base-RL & 11/50 (22\%) & [12.8, 35.2] & \(204.1\pm21.3\) \\
Ours (full) & \textbf{30/50 (60\%)} & \textbf{[46.2, 72.4]} & \(\mathbf{140.7\pm15.9}\) \\
\bottomrule
\end{tabular}}
\end{table}

\subsubsection{Height Adaptation}

\begin{figure}[t]
    \centering
    \includegraphics[width=\columnwidth, height=4cm]{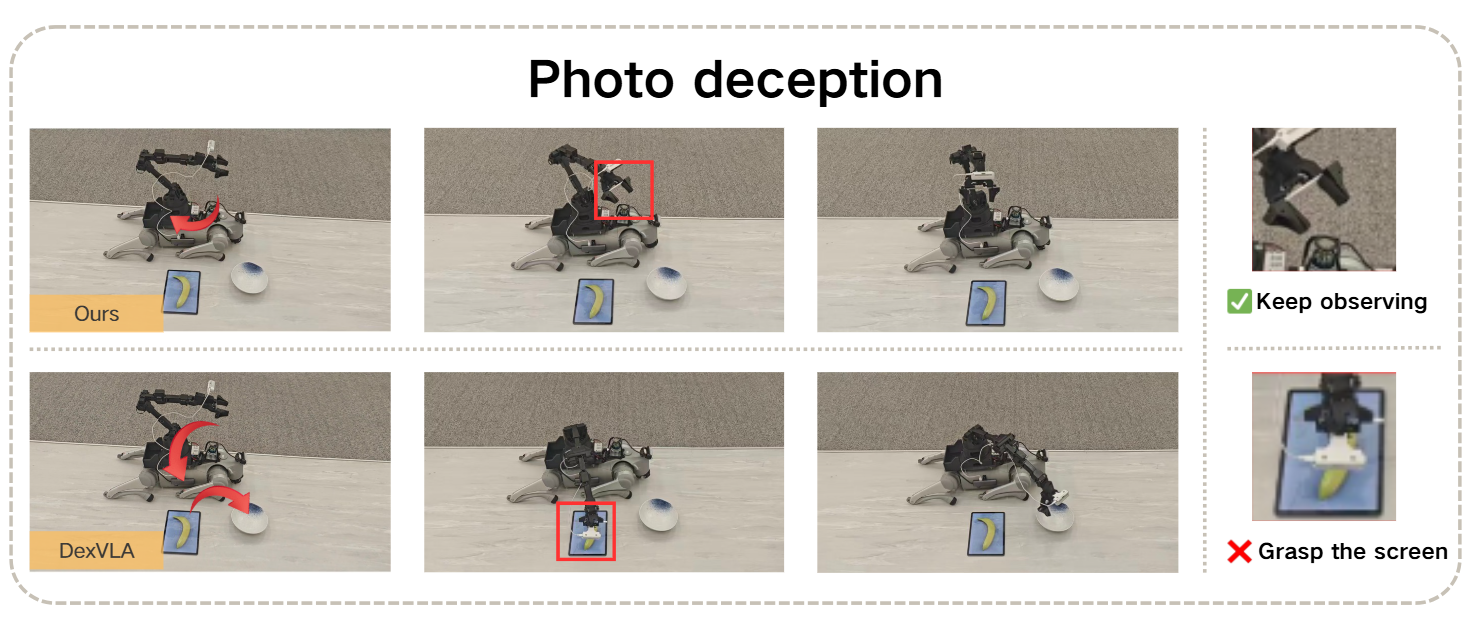}
    \vspace{-20pt}
    \caption{Photo-deception setup. The robot should manipulate the real object while rejecting a photo-realistic distractor on a flat screen.}
    \Description{Qualitative comparison between successful rejection of a banana image on a tablet and a baseline that attempts to grasp the screen.}
    \label{fig:photo}
\end{figure}

\begin{figure}[t]
    \centering
    \includegraphics[width=\columnwidth, height=4cm]{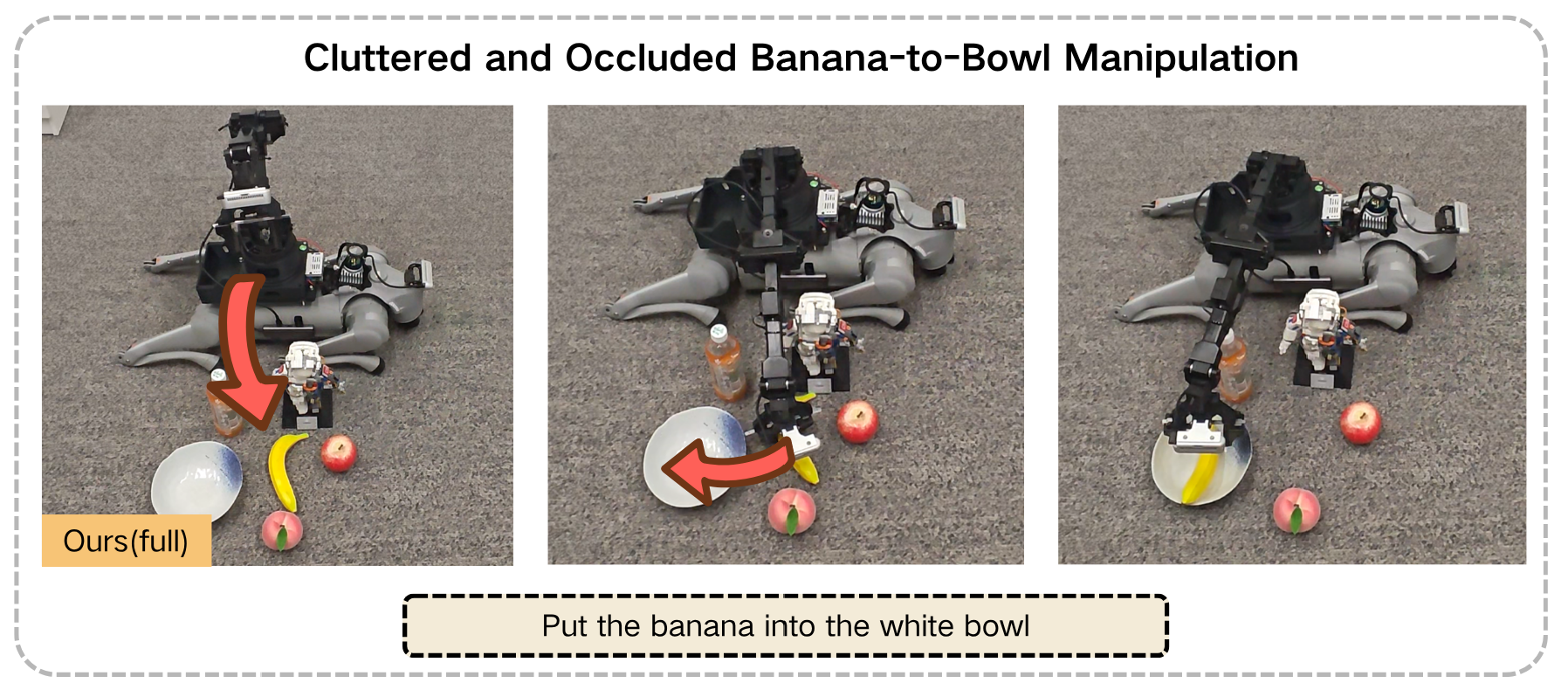}
    \vspace{-20pt}
    \caption{Cluttered banana-to-bowl benchmark with physical distractors and strong target occlusion.}
    \Description{A cluttered tabletop scene containing a banana, bowl, LEGO astronaut, fruit, and bottle. The sequence illustrates active-view grounding followed by grasp-and-place execution.}
    \label{fig:clutter}
\end{figure}

Figure~\ref{fig:height} shows the height-shift settings, while Table~\ref{tab:height_result} reports the corresponding results. The full system remains between 75\% and 80\% success, while the variant without Base-RL fails when the initial stance is retained.

\begin{table}[t]
\caption{Height adaptation across platform height offsets (cm).}
\label{tab:height_result}
\centering
\begin{tabular}{lccc}
\toprule
Method & \(0\!\to\!30\) & \(0\!\to\!60\) & \(0\!\to\!75\) \\
\midrule
DexVLA & 48\% & 33\% & 23\% \\
PointVLA & 58\% & 46\% & 35\% \\
Ours w/o Base-RL & 0\% & 0\% & 0\% \\
Ours (full) & \textbf{80\%} & \textbf{78\%} & \textbf{75\%} \\
\bottomrule
\end{tabular}
\end{table}

\subsubsection{Photo Deception}
Table~\ref{tab:photo_result} replaces the real target with a photo-realistic distractor. RGB-only DexVLA and the single-view ablation exhibit high false-grasp rates, while PointVLA and the full method reject the flat-screen distractor. The full model also improves real-object success, indicating more stable localization beyond distractor rejection.

\begin{table}[t]
\caption{Robustness to photo deception.}
\vspace{-5pt}
\label{tab:photo_result}
\centering
\begin{tabular}{lcc}
\toprule
Method & Real success \(\uparrow\) & Photo false-grasp \(\downarrow\) \\
\midrule
DexVLA & 78\% & 76\% \\
PointVLA & 80\% & 0\% \\
Ours Single-View & 76\% & 70\% \\
Ours (full) & \textbf{88\%} & \textbf{0\%} \\
\bottomrule
\end{tabular}
\end{table}

\subsubsection{Cluttered and Occluded Manipulation}
The clutter benchmark stresses both target visibility and nearby obstacle geometry. Specifically, Table~\ref{tab:clutter_result} illustrates that active multi-view Semantic-3DGS raises success from 52\% for the single-view variant to 74\%, while improving the collision-free rate from 70\% to 88\% and reducing false grasps from 18\% to 6\%. Relative to PointVLA, the full system improves success by 28 percentage points.

\begin{table*}[!t]
\caption{Cluttered banana-to-bowl benchmark over 50 trials per method. CI denotes a 95\% Wilson interval.}
\vspace{-8pt}
\label{tab:clutter_result}
\centering
\begin{tabular}{lccccc}
\toprule
Method & Success & 95\% CI & Collision-free & False grasp & Avg. time (s) \\
\midrule
DexVLA & 13/50 (26\%) & [15.9, 39.6] & 22/50 (44\%) & 16/50 (32\%) & \(27.9\pm3.1\) \\
PointVLA & 23/50 (46\%) & [33.0, 59.6] & 31/50 (62\%) & 10/50 (20\%) & \(30.7\pm3.4\) \\
Ours Single-View & 26/50 (52\%) & [38.5, 65.2] & 35/50 (70\%) & 9/50 (18\%) & \(29.5\pm3.2\) \\
Ours (full) & \textbf{37/50 (74\%)} & \textbf{[60.4, 84.1]} & \textbf{44/50 (88\%)} & \textbf{3/50 (6\%)} & \(33.2\pm3.6\) \\
\bottomrule
\end{tabular}
\end{table*}

\paragraph{Component ablations.}
Separate 30-trial clutter ablations isolate the main representation and conditioning choices. A VGGT point-map variant reaches 58\% success. Removing CLIP/DINO semantic features yields 60\%, and removing the obstacle-occupancy cue yields 65\%. All-block semantic injection reaches 68\%, compared with 74\% for the full system. Collision-free execution and false-grasp rate degrade in the same ablations, indicating that the gains do not arise from active sensing or base repositioning alone.

\subsubsection{Late-block Injection and Runtime}
\begin{figure}[t]
    \centering
    \includegraphics[width=\columnwidth, height=5cm]{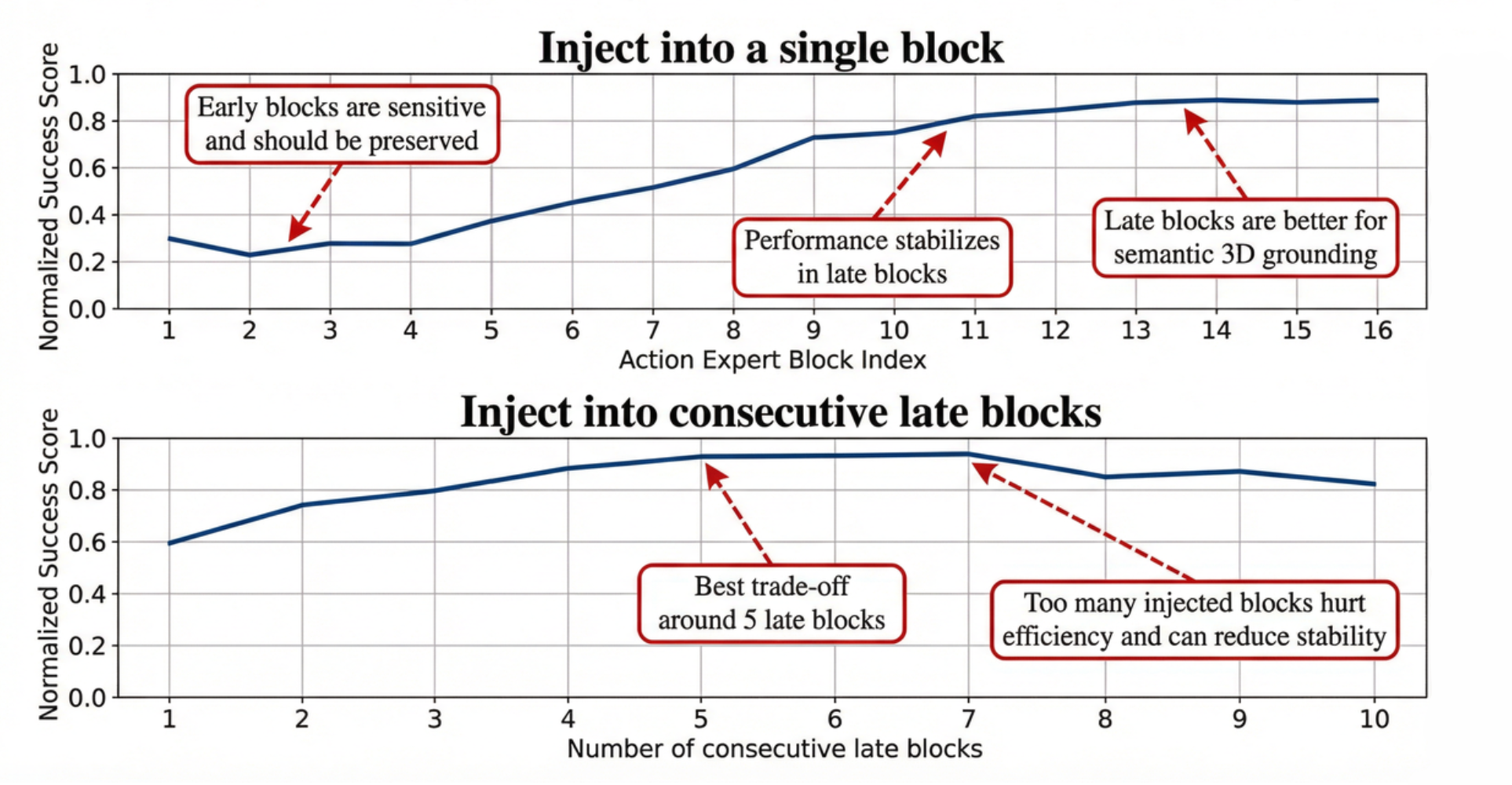}
    \vspace{-20pt}
    \caption{Block sensitivity analysis for semantic-3DGS injection: late-block conditioning provides a better success--efficiency trade-off than injecting semantic cues into earlier or broader portions of the action expert.}
    \Description{A plot comparing semantic-injection locations across the action expert. Performance and latency jointly favor conditioning the final blocks rather than earlier or all blocks.}
    \label{fig:semantic_block}
\end{figure}

Late-block injection provides a better success/latency trade-off than modifying all action-expert blocks (Table~\ref{tab:block_result}). Figure~\ref{fig:semantic_block} further shows the block-sensitivity trend: earlier or broader intervention disrupts the pretrained action prior more strongly than conditioning only the late blocks. Runtime is reported at three levels: one-time grounding latency, per-action-chunk online latency, and independently measured full-task wall-clock time. We keep these measurements separate: the one-time grounding costs and per-chunk online latencies are not arithmetically summed to estimate task duration. Full-task wall-clock time is measured independently from the start of active sensing to completion of placement and already includes repeated action-chunk inference and communication, physical arm/base motion, grasping, transfer, placement, and settling.

\begin{table}[!t]
\caption{Late-block versus all-block semantic injection.}
\vspace{-10pt}
\label{tab:block_result}
\centering
\begin{tabular}{lcc}
\toprule
Variant & Avg. success & Chunk latency \\
\midrule
All-block & 75\% & 175 ms \\
Late-block (5) & \textbf{82\%} & \textbf{80 ms} \\
\bottomrule
\end{tabular}
\end{table}

\begin{table}[!t]
\caption{Runtime profile. One-time stages execute before manipulation; online values are measured per action chunk.}
\vspace{-10pt}
\label{tab:runtime}
\centering
\begin{tabular}{lc}
\toprule
Component & Mean latency \\
\midrule
Four-view wrist motion + capture & 16.00 s \\
VGGT pose/depth initialization & 0.62 s \\
Semantic-3DGS feature update & 1.21 s \\
Semantic rendering + localization & 0.34 s \\
VLA action-chunk inference & 0.08 s / chunk \\
ROS/WiFi communication & 0.05 s / chunk \\
\midrule
Full clutter task (ours) & \(33.2\pm3.6\) s \\
\bottomrule
\end{tabular}
\end{table}

Active multi-view sensing adds about 3.7\,s to the clutter-task wall-clock time (33.2\,s versus 29.5\,s for Single-View), while improving success by 22 points and collision-free execution by 18 points. This quantifies the robustness--latency trade-off of the up-front grounding stage rather than implying real-time 3D reconstruction inside the manipulation servo loop.

\paragraph{Scope and limitations.}
Our system targets quasi-static local household manipulation where a short active grounding stage is acceptable. Semantic-3DGS construction and VLA inference currently run on an off-board GPU, and the local representation is built before manipulation or refreshed after grounding failure rather than updated inside the low-level servo loop. We therefore do not claim fast dynamic interaction or zero-shot acquisition of arbitrary manipulation skills; the evaluated setting is open-vocabulary target grounding with few-shot embodiment-specific manipulation.

\section{Conclusion}
We presented an embodied multimodal grounding framework for open-vocabulary target grounding with few-shot mobile manipulation of robots. A refreshable Semantic-3DGS provides a shared interface across active sensing, language-conditioned localization, obstacle-aware geometry, reachability-aware stance preparation, and late-block VLA conditioning. Expanded real-robot evaluations show improved robustness under long-horizon execution, height variation, photo-realistic distractors, and heavy clutter. Future research will investigate lighter onboard representations, adaptive view planning, and broader unseen-object generalization.

\begin{acks}
This work was supported by the National Natural Science Foundation of China under Grants 62303389 and 62373289. It was also supported by National Key Research and Development Program of China under Grant 2025YFB4713002. Additional support came from Guangdong Scientific Research Platform and Project Scheme under Grant 2024KTSCX039, Guangzhou-HKUST(GZ) Joint Funding Program under Grant
2024A03J0618, the Youth Talent Support Program of Guangdong Provincial Association for Science and Technology under Grant SKXRC2025463 and Guangdong Provincial Key Lab of Integrated Communication, Sensing and Computation for Ubiquitous Internet of Things (grant number 2023B1212010007).
\end{acks}

\bibliographystyle{ACM-Reference-Format}
\bibliography{acmmm26}

\end{document}